\documentclass{article}

\usepackage{microtype}
\usepackage{graphicx}
\usepackage{subfigure}
\usepackage{booktabs} % for professional tables

\usepackage{hyperref}

\usepackage[preprint]{icml2024}

\usepackage{amsmath}
\usepackage{amssymb}
\usepackage{mathtools}
\usepackage{amsthm}

\usepackage[capitalize,noabbrev]{cleveref}

\theoremstyle{plain}

\theoremstyle{definition}

\theoremstyle{remark}

\usepackage{amsmath}
\usepackage{amssymb}
\usepackage{makecell}
\usepackage{wrapfig}

\DeclareMathOperator*{\argmin}{arg\,min}

\newcommand{\Ac}{\mathcal{A}}
\newcommand{\Cc}{\mathcal{C}}

\newcommand{\Rc}{\mathcal{R}}
\newcommand{\Sc}{\mathcal{S}}
\newcommand{\Tc}{\mathcal{T}}

\newcommand{\zvec}{\boldsymbol{z}}
\newcommand{\muvec}{\boldsymbol{\mu}}
\newcommand{\sigmavec}{\boldsymbol{\sigma}}

\usepackage[textsize=tiny]{todonotes}

\icmltitlerunning{Skill-based Safe Reinforcement Learning with Risk Planning}

\begin{document}

\twocolumn[
\icmltitle{Skill-based Safe Reinforcement Learning with Risk Planning}



\icmlsetsymbol{equal}{*}

\begin{icmlauthorlist}
\icmlauthor{Hanping Zhang}{inst}
\icmlauthor{Yuhong Guo }{inst}
\end{icmlauthorlist}

\icmlaffiliation{inst}{School of Computer Science, Carleton University, Ottawa, Canada}

\icmlcorrespondingauthor{ }{jagzhang@cmail.carleton.ca}
\icmlcorrespondingauthor{ }{yuhong.guo@carleton.ca}

\icmlkeywords{Machine Learning, ICML}

\vskip 0.3in
]



\printAffiliationsAndNotice{\icmlEqualContribution} % otherwise use the standard text.

\begin{abstract}
Safe Reinforcement Learning (Safe RL) aims to ensure safety 
when an RL agent conducts learning by interacting with real-world environments 
where improper actions can induce high costs or lead to severe consequences.
In this paper, we propose a novel 
Safe Skill Planning (SSkP) 
approach to enhance effective safe RL by exploiting auxiliary offline demonstration data. 
SSkP involves a two-stage process. 
First, we employ PU learning to learn a skill risk predictor 
from the offline demonstration data. 
Then, based on the learned skill risk predictor,
we develop a novel 
risk planning process
to enhance online safe RL and learn a risk-averse safe policy efficiently through 
interactions with the online RL environment, 
while simultaneously adapting the skill risk predictor to the environment. 
We conduct experiments in several benchmark robotic simulation environments.
The experimental results demonstrate that the proposed approach consistently outperforms 
previous state-of-the-art safe RL methods.  
\end{abstract}
\section{Introduction}
Reinforcement Learning (RL) empowers the development of intelligent agents and the training of decision systems,
making it highly suitable for real-world applications. As RL continues to find broader use in real-world scenarios,
concerns regarding the safety of RL systems have become more noticeable. These safety concerns have been particularly
highlighted in human-centric domains, such as autonomous driving~\cite{wen2020safe}, 
helicopter manipulation~\cite{koppejan2011neuroevolutionary},
and human-related robotic environments~\cite{brunke2021safe},
where significant risks can be associated with taking improper actions,
leading to severe consequences.

Safe Reinforcement Learning (Safe RL) focuses on the development of 
RL systems while adhering
to predefined safety constraints~\cite{garcia2015comprehensive}
and reducing the associated risk. 
In Safe RL, in addition to optimizing a reward
function~\cite{sutton2018reinforcement}, an additional cost is often assigned to evaluate the safety of actions taken by the
RL agent; the RL agent aims to maximize the reward signal 
while ensuring a low cost
\cite{altman1999constrained,hans2008safe}.
Conventional Safe RL methods aim to 
maximize cumulative rewards through interactions with online
environments~\cite{achiam2017constrained,tessler2018reward,thomas2021safe},
which often incur nontrivial costs in the learning process. 
More recently, researchers have recognized
the value of learning from offline data, a practice that avoids potential damage to online physical
environments~\cite{xu2022constraints,liu2023datasets}.
Reinforcement Learning from Demonstration (LfD)
seeks to accelerate RL training by initially pre-training the RL agent 
using an offline dataset of demonstrations,
which has demonstrated effective performance for standard RL tasks
\cite{argall2009survey,brys2015reinforcement}. 
Recent research has started to exploit 
the potential of LfD in the context of Safe RL, 
aiming to incorporate the safety-related information from the demonstration data to improve
the training of safe policies in online environments
\cite{thananjeyan2021recovery}.
Our research endeavors to further advance safe RL in this intriguing direction.

In this paper, we introduce a novel Safe Skill Planning (SSkP) approach to 
enhance effective safe online RL by exploiting the offline demonstration data. 
Skill learning is a commonly used technique for LfD, allowing the RL agent to 
learn high-level representations of action sequences from offline demonstrations~\cite{pertsch2021accelerating}.
In SSkP, we first employ a skill model to capture the high level behaviour patterns in the offline demonstrations
as latent skills, and learn a skill risk predictor through Positive-Unlabeled (PU) learning 
on the demonstration data.
The skill risk predictor estimates the level of risk associated with executing a skill-based 
action sequence in a given state.
Subsequently, we use the skill risk predictor 
to evaluate the safety of an RL agent's exploration behaviors (skills),
and develop a novel 
risk planning process 
to enhance safe exploration 
and facilitate the efficient learning of a safe policy 
through interactions with online RL environments,
while adapting the skill risk predictor to 
these online environments in real-time.
We conduct experiments
in various robotic simulation environments~\cite{thomas2021safe} built on
Mujoco~\cite{todorov2012mujoco}.
The experimental results 
demonstrate that our proposed approach 
produces superior performance over 
several state-of-the-art safe RL methods, such as 
Recovery RL~\cite{thananjeyan2021recovery}, 
CPQ~\cite{xu2022constraints} and SMBPO~\cite{thomas2021safe}. 
Our main contributions can be summarized as follows:
\begin{itemize}
\item 
We propose an innovative skill risk prediction methodology for  
extracting safe decision evaluation information from offline demonstration data 
and facilitating safe RL in online environments. 
\item
We devise a novel risk planning process aimed at generating safer skill decisions
by leveraging skill risk prediction,
thereby enhancing safe exploration and learning in online RL environments. 
\item
The proposed method SSkP demonstrates superior performance over the state-of-the-art safe RL methods. 
\end{itemize}

%%%%%%%%%%%%%%%%%%%%%%%%%%%%%%%%%%%%%%
\section{Related Works}
\paragraph{\bf Safe RL}
Safe Reinforcement Learning (Safe RL) is the study of optimizing decision-making for RL systems while
ensuring compliance with safety constraints. 
It aims to strike a balance between exploration for learning and
the avoidance of actions that could result in harmful or undesirable outcomes \citep{garcia2015comprehensive}.
\citet{altman1999constrained} first introduced the formulation of Constrained Markov Decision Processes (CMDPs) to frame the Safe RL problem.
Subsequent research in \citep{hans2008safe} introduced strict constraints that prohibit safety violations within a single
exploration trajectory. 
\citet{thomas2021safe} developed a Safe Model-Based Policy Optimization (SMBPO) method, aiming to learn a precise
transition model that prevents unsafe states during exploration by penalizing unsafe trajectories.
Recent studies have highlighted the significance of incorporating offline data into Safe RL. 
\citet{xu2022constraints} introduced Constrained Penalized Q-learning (CPQ), which employs a cost critic to learn 
constraint values during exploration. They further penalize the Bellman operator in policy training to stop the update of the policy for potentially unsafe states.
In another endeavor, \citet{thananjeyan2020safety} proposed the Safety Augmented Value Estimation from Demonstrations
(SAVED) approach, facilitating the learning of a safety density model from offline demonstration data. They utilize the
cross-entropy method \citep{botev2013cross} for planning safe exploration, balancing task-driven exploration with 
cost-driven constrained exploration.
Their more recent work introduced a Recovery RL approach \citep{thananjeyan2021recovery}, learning a recovery policy
from offline demonstration data. This method ensures a recovery policy's safety by leveraging demonstration data, while
also learning a recovery set to evaluate state safety. During online training, a task policy is learned when states are
deemed safe, switching to the recovery policy when the RL agent encounters potentially unsafe situations.
\paragraph{\bf Skill-based RL}
Reinforcement Learning from Demonstration (LfD), also known as Imitation Learning, focuses on enhancing online RL
training by leveraging an expert demonstration dataset \citep{argall2009survey, brys2015reinforcement}.
\citet{thrun1994finding} 
introduced skill learning to LfD, enabling RL agents to learn reusable high-level skills
from action sequences within offline demonstration data. In more recent research,
\citet{pertsch2021accelerating} presented the SPiRL framework, which leverages deep latent models to
learn skill representations. The policy is trained using the skill model in conjunction with a variant of
Soft Actor-Critic (SAC) \citep{haarnoja2018soft} to accelerate RL in downstream tasks.
Furthermore, recent work has demonstrated the integration of skill learning into {\em offline} safe RL \citep{slack2022safer},
which learns a safety variable posterior from offline demonstration data and subsequently enhances online
safe policy training. 
\paragraph{\bf Positive-Unlabeled Learning}
In contrast to traditional supervised learning that relies on labeled positive and negative examples,
Positive-Unlabeled (PU) learning addresses scenarios where data cannot be strictly categorized as positive or negative.
Notably, \citet{du2014analysis,du2015convex}'s previous work introduced an unbiased estimation of the true negative
loss, making PU learning feasible. \citet{jain2016estimating} and \citet{christoffel2016class} extended this research by
enhancing the accuracy of practical PU classifier training through positive class prior estimation.
\citet{kiryo2017positive} proposed a large-scale PU learning approach that addresses overfitting by introducing 
non-negative constraints and a relaxed slack variable.
In recent developments, \citet{xu2021positive} applied PU learning to Generative Adversarial
Imitation Learning (GAIL) \citep{ho2016generative} in RL,
which learns an optimized reward function from the expert demonstration dataset to 
improve RL performance in offline training.

%%%%%%%%%%%%%%%%%%%%%%%%%%%%%%%%%%%%%%%%%%%%%%%%%%%%%%%%%%
%
\section{Problem Setting}
The safe RL
problem is typically framed as a Constrained Markov Decision Process (CMDP)~\cite{altman1999constrained}, denoted as
$M=(\Sc, \Ac, \Tc, \Rc, \Cc, \gamma)$, where $\Sc$ represents the state space, $\Ac$ is the action space,
$\Tc: \Sc\times \Ac\to \Sc$ defines the transition dynamics, $\Rc: \Sc\times \Ac\to \mathbb{R}$ is the reward function,
and $\gamma\in (0, 1)$ is the discount factor.
The additional cost function $\Cc: \Sc\times \Ac\to \mathbb{R}$ is 
introduced to account for safety violations during RL exploration. 
Hence an exploration trajectory within CMDP can be expressed
as $\tau=(s_0, a_0, r_0, c_0, \ldots, s_t, a_t, r_t, c_t, \ldots, s_{|\tau|+1})$. 
We adopt the strict setting that
the safe RL agent will terminate a trajectory 
when encountering safety violation and inducing a nonzero cost ($c_t>0$)~\cite{hans2008safe}.
The goal of safe RL is to efficiently learn a good policy $\pi$ 
that maximizes expected discounted cumulative reward while incurring minimal costs.

To facilitate safe RL in online environments, 
we presume the availability of a small demonstration dataset, denoted as $\mathcal{D}_d$, 
which provides prior information regarding safety violations 
during exploration: $\mathcal{D}_d=\{\ldots,(\cdots, s_t, a_t, c_t,\cdots),\ldots\}$.
The demonstration data can be gathered by either human experts or a trained safe RL agent
\cite{thananjeyan2021recovery}. 
A method that can effectively exploit such demonstration data is expected to 
accelerate safe RL in online environments with smaller costs. 

%%%%%%%%%%%%%%%%%%%%%%%%%%%%%%%%%%%%%%%%%%%%%%%%%%%%%%%%%%
\begin{figure*}[t]
\centering
\includegraphics[width=.85\textwidth]{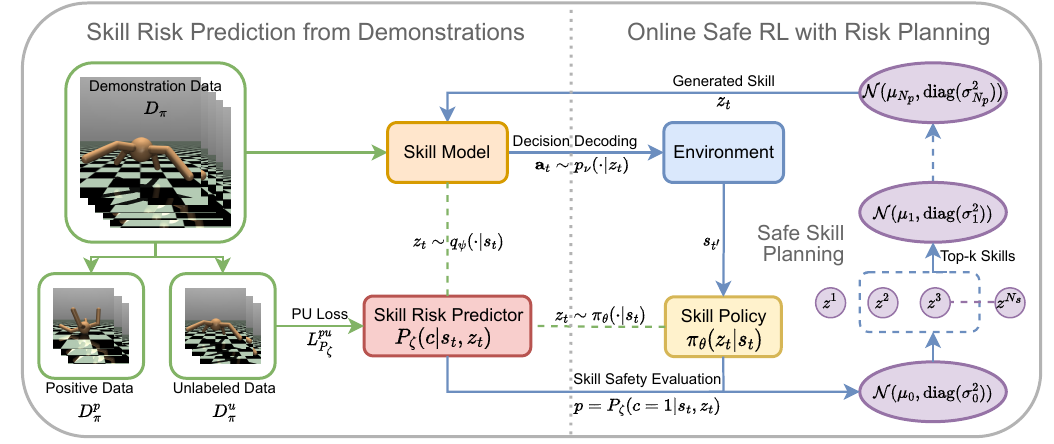}
\caption{
The framework of the proposed method, SSkP, 
which learns a skill risk predictor from the offline demonstration data
and then deploys it to enhance online safe RL through risk planning. 
During the skill risk predictor learning stage, SSkP assembles PU data 
and trains a decision risk predictor $P_\zeta(c|s_t,z_t)$ based on a skill model,
which produces skill prior $q_\psi(\cdot|s_t)$ and skill decoder $p_\nu(\textbf{a}_t|z_t)$.
In the online safe policy learning stage, 
a risk planning process is deployed to generate and choose safer skill decisions 
based on the skill risk predictor $P_\zeta(c|s_t,z_t)$. 
	The generated skill $z_t$ is decoded by the skill decoder $p_\nu(\textbf{a}_t|z_t)$
into an action sequence $\textbf{a}_t$ to interact with the online environment. 
	Rewards are collected from these online interactions
to learn the safe skill policy $\pi_\theta(z_t|s_t)$. 
}
\label{fig:architecture}
\end{figure*}
%
%%%%%%%%%%%%%%%%%%%%%%%%%%%%%%%%%%%%%%%%%%%%%%%%%%%%%%%%%%
%
\section{Method}
The main framework of the proposed 
Safe Skill Planning (SSkP) 
approach is presented in Figure~\ref{fig:architecture},
which has two stages: 
skill-risk predictor learning and 
safe RL with risk planning. 
Towards the goal of facilitating efficient safe RL, 
SSkP first exploits the prior demonstration data to extract reusable high-level skills
and learn a skill risk predictor through PU learning.  
Then by devising a risk planning process based on
the skill risk predictor, the online RL agent is guided to
pursue risk-averse explorations 
and efficiently learn a skill policy in online environments
that can maximize the expected reward
with minimal costs. 
We further elaborate these two stages in the following subsections.

\subsection{Skill Risk Prediction from Demonstrations}
\label{sec:risk_prediction}
Conventional safe RL methods entail the learning of a safe policy through direct interaction with
the online environment, 
which often incur considerable costs in the exploration based learning process.
Learning from demonstration (LfD)
offers a means to accelerate the online RL process 
and reduce the cost by pre-training on an offline demonstration dataset. 
This pre-training phase
is more efficient in terms of time and cost compared to the resource-intensive online environment.
Skill-based learning stands as a prominent approach in LfD~\cite{pertsch2021accelerating}.
It learns reusable skills as generalizable high level representations of action sequences from 
offline demonstrations, which can be used to guide the RL agent to explore 
in a safe manner for downstream online tasks. 
Inspired by the principles of LfD, 
we aim to extract skill-based safety-related insights 
from the demonstration dataset $\mathcal{D}_d$,
which can be utilized to assess the safety of reinforcement decisions
and enhance the ensuing online safe RL. 
In particular, we propose to learn a skill risk predictor  $P_\zeta(c|s_t, \zvec_t)$ from the demonstration data
that can evaluate the safety of a skill-based decision, $\zvec_t$,
on a given state $s_t$.

To support skill-based learning, 
we first adopt the deep skill model from a previous work~\cite{pertsch2021accelerating} 
to learn skills as latent representations of observed action sequences. 
This skill model consists of three key components: a skill encoder network
$q_\mu(\zvec_t|{\bf a}_t)$, responsible for encoding an action sequence 
${\bf a}_t=\{a_t, ..., a_{t+H-1}\}$ with length $H$
into a high-level skill $\zvec_t$;
a skill decoder network $p_\nu({\bf a}_t|\zvec_t)$, which decodes the skill $\zvec_t$ back into the action sequence ${\bf a}_t$;
and a skill prior network $q_\psi(\zvec_t|s_t)$, which generates the skill decision for a given state $s_t$.
After being trained on the demonstration data $\mathcal{D}_d$, 
the components of the skill model can be deployed to facilitate subsequent learning processes.  

\subsubsection{Learning Skill Risk Predictor via PU Learning}
The demonstration data provides valuable insights for safe exploration of the environment. 
However, estimating risk predictors for skill-based behaviors 
in the context of safe exploration poses a persistent challenge 
due to two primary reasons. 
First, the demonstration data, whether collected by a human expert or a fully trained safe RL agent, 
often contain very limited actual examples of safety violations, 
due to the finite trajectory lengths and limited skill horizons. 
Second, while a decision made in a given state may not result in immediate safety violations, 
it could lead to a close proximity to safety violations. 
Treating such decisions as strictly safe examples can be problematic.
To tackle these issues, we propose the utilization of Positive-Unlabeled (PU) learning, 
a technique that can bypass the strict differentiation of safe decisions from unsafe ones
as well as alleviate the scarcity of unsafe examples.

Specifically, we collect the positive and unlabeled decision examples for PU learning as follows. 
At a timestep $t$, if the current trajectory $\tau$ actually 
encounters a safety violation within the next $H$ steps 
when the RL agent is projected to select skill $\zvec_t$ at state $s_t$, then 
we collect such state-skill pair $(s_t,\zvec_t)$ as {\em positive} unsafe examples.
Conversely, all other state-skill decision pairs that do not lead to immediate risks are collected as {\em unlabeled} examples. 
For states near the termination of trajectories, 
the corresponding action sequences have lengths that are insufficient (less than the horizon $H$) to encode skills.
We hence utilize the skill prior network $q_\psi(\zvec_t|s_t)$ from the skill model 
to produce the skill decision $\zvec_t$ for each given state $s_t$ in the demonstration data, instead of using the encoder.

Let $D^p={(s_t^p, \zvec_t^p)}$ represent the set of positive examples of state-skill decision pairs, 
and $D^u={(s_t^u, \zvec_t^u)}$ represent the unlabeled set. 
We learn the skill risk predictor $P_\zeta(c|s_t, \zvec_t)$ 
as a binary classifier parameterized with $\zeta$, 
measuring the probability of selecting skill $\zvec_t$ at state $s_t$ leading to a safety violation
with risk $c>0$.
We compute the true positive loss on the PU training data as the negative mean log-likelihood of
the positive examples in $D^p$: 
\begin{align}
	{L}_{P_\zeta}^1(D^p) &= -{\mathbb{E}}_{(s,\zvec)\sim D^p} [\log(P_\zeta(c=1|s,\zvec))],
\end{align}
while the difficulty lies in computing the true negative loss without confirmed negative examples. 
To bypass this problem, unbiased estimation of the true negative loss
using PU data has been developed in the literature~\cite{du2015convex,du2014analysis}:
\begin{align}
	{L}_{P_\zeta}^0(D^u\cup D^p) = 
	{L}_{P_\zeta}^0(D^u) - \lambda {L}_{P_\zeta}^0(D^p)
\end{align}
where $\lambda$ represents the positive class prior, which can be estimated using positive and unlabeled data
\cite{jain2016estimating,christoffel2016class};
${L}_{P_\zeta}^0(D)$ denotes the negative mean log-likelihood of
the given data $D$ being negative, such that:
\begin{align}
	{L}_{P_\zeta}^0(D) &= -{\mathbb{E}}_{(s,\zvec)\sim D} [\log(1-P_\zeta(c=1|s,\zvec))].
\end{align}
To further improve the estimation of the true negative loss, 
in the recent PU learning literature, Kiryo et al.~\cite{kiryo2017positive} introduce an
additional constraint to the estimation of 
${L}_{P_\zeta}^0(D^u\cup D^p)$,  
ensuring that the loss remains non-negative: 
${L}_{P_\zeta}^0(D^u) - \lambda {L}_{P_\zeta}^0(D^p) \geq 0$.
To provide tolerance and reduce the risk of overfitting, 
a non-negative slack variable $\xi\geq 0$ is also introduced to relax
the constraint, which leads to the following PU loss we adopted
for training our skill risk predictor: 
\begin{align}
	{L}_{P_\zeta}^{pu}(D^p, D^u) =&
	\lambda {L}_{P_\zeta}^1(D^p) + 
	\nonumber\\ 	
	& \max(-\xi, {L}_{P_\zeta}^0(D^u) - \lambda {L}_{P_\zeta}^0(D^p))
\label{eqa:pu_loss}
\end{align}
By minimizing this PU loss on the demonstration data, we obtain 
a pre-trained skill risk predictor $P_\zeta(c|s_t, \zvec_t)$,
which will be deployed in the online RL stage to screen the skill decisions
and accelerate safe policy learning.

%%%%%%%%%%%%%%%%%%%%%%
\subsection{Online Safe RL with Risk Planning}
\label{sec:online}
In the online safe policy learning stage, 
our objective is to facilitate the learning of a safe policy by leveraging the
safe skill knowledge learned from the offline demonstration data,
encoded by the skill prior network $q_\psi(\cdot|s_t)$, 
the decoder network $p_\nu(\cdot|\zvec_t)$, and, in particular, the skill risk predictor $P_\zeta(c|s_t, \zvec_t)$. 

\subsubsection{Risk Planning}
\label{sec:riskplan}
%
%%%%%%%%%%%%%%%%%
\begin{algorithm}[t]
\caption{Risk Planning}
\label{algorithm:risk_planning}
\textbf{Initialize:}
$(\muvec_0,\sigmavec_0^{2})\leftarrow q_\psi(\cdot|s_t)$ \\
\textbf{Procedure:}
\begin{algorithmic}[1]
\FOR{$i=1,2,...,N_p$}
	\STATE Sample skills $\{\zvec^j\}_{j=1}^{N_s}$ from $\mathcal{N}(\muvec_{i-1},\text{diag}(\sigmavec_{i-1}^2))$
    \STATE Calculate $p_j=P_\zeta(c=1|s_t,\zvec^j)$ for $N_s$ skills
	\STATE Compute $(\muvec_i, \sigmavec_i^2)$ using the selected top-k skills with lowest risk predictions in $\{p_j\}_{j=1}^{N_s}$ 
\ENDFOR
\STATE Sample skill ${\zvec}_t$ from $\mathcal{N}(\muvec_{N_p},\text{diag}(\sigmavec_{N_p}^2))$
\end{algorithmic}
\end{algorithm}
%
%%%%%%%%%%%%%%%%%
%
The pre-trained skill risk predictor $P_\zeta(c|s_t, \zvec_t)$ 
encodes safe decision evaluation information extracted from the demonstration data,
providing an essential capacity for pre-assessing the safety of potential skill-based decisions 
before executing them in online environments. 
Specifically, $P_\zeta(c=1|s_t, \zvec_t)$ can 
quantify the likelihood that the RL agent will encounter safety violation
by following the action sequence encoded by skill $\zvec_t$ at state $s_t$. 
We have, therefore, developed a heuristic 
risk planning process that leverages the skill risk predictor 
to choose safer skill decisions to follow. 
This process is expected to reduce the potential for encountering safety violations 
and enhance the safety of online RL learning.

Specifically, we evaluate and choose skill-based decisions 
at a given state $s_t$ from an iteratively self-enhanced Gaussian distribution $\mathcal{N}(\muvec, \text{diag}(\sigmavec^2))$
that has a diagonal covariance matrix.  
At the start, we sample $N_s$ skills from the current safe policy 
function $\pi_\theta(\cdot|s_t)$ at state $s_t$ 
such that 
$\{\zvec^j\sim\pi_\theta(\cdot|s_t),\ j=1\cdots N_s\}$,
and use these skill vectors to calculate the mean and covariance 
of an initial Gaussian distribution  
$\mathcal{N}(\muvec_0, \text{diag}(\sigmavec_0^2))$.
Then in each $i$-th iteration, we sample $N_s$ skills
$\mathcal{Z}=\{\zvec^j\}_{j=1}^{N_s}$
from the current Gaussian distribution
$\mathcal{N}(\muvec_{i-1}, \text{diag}(\sigmavec_{i-1}^2))$ 
and evaluate their safety using the skill risk predictor
$p_j = P_\zeta(c|s_t,\zvec^j)$. 
We choose the top-k safe skills $\mathcal{Z}_k$ with the lowest predicted risk probabilities 
from $\mathcal{Z}$ 
to update the Gaussian distribution for the next iteration: 
\begin{align}
	\label{eq:updatemu}
	\muvec_{i} &= \frac{1}{k}\sum\nolimits_{\zvec\in\mathcal{Z}_k}\zvec, \\
	\sigmavec_{i}^2 &= \frac{1}{k}\sum\nolimits_{\zvec\in\mathcal{Z}_k}
			\text{diag}\left((\zvec - \muvec_{i})(\zvec - \muvec_{i})^\top\right)
	\label{eq:updatesigma}
\end{align}
After a total number of $N_p$ iterations, an optimized skill decision ${\zvec}_t$ 
with low predicted risk is sampled from the final refined distribution
$\mathcal{N}(\muvec_{N_p}, \text{diag}(\sigmavec_{N_p}^2))$.
The procedure of this planning process is also summarized in
Algorithm~\ref{algorithm:risk_planning}.
This risk planning procedure
is essentially a cross-entropy method (CEM) 
\cite{botev2013cross,rubinstein1997optimization},
specifically employed in this context
as a zeroth-order solver to tackle the non-convex optimization problem \cite{amos2020differentiable} 
of $\argmin_{\zvec} P_\zeta(c=1|s_t, \zvec)$, 
facilitating effective selection of safe skills based on the skill risk predictor. 
By gradually adjusting the Gaussian distribution towards safer decision skill regions, 
we expect to reliably identify a safe skill after a sufficient number of iterations.

\subsubsection{Online Safe Policy Learning}
By utilizing the pre-learned skill knowledge and the proposed risk planning process, 
we aim to efficiently learn a skill-based safe policy network $\pi_\theta(\zvec|s)$ 
through iterative interactions with an online RL environment,
which maximizes the expected discounted reward while minimizing the costs
incurred by safety violations. 
Specifically, 
at the current state $s_t$, 
we first select an optimized skill, ${\zvec}_t$, using the risk planning process. 
This skill, ${\zvec}_t$, is then decoded into an action sequence, ${\bf a}_t = a_{t:t+H-1}$, using the skill decoder
$p_\nu(\cdot|{\zvec}_t)$. 
The RL agent interacts with the online environment 
to reach next state $s_{t'}$ by taking this sequence of actions,
adhering to the behavior patterns of the pre-learned skills. 
During the interaction process, the RL agent collects cumulative reward signals 
$\tilde{r}_t=\sum_t^{t'-1}r_t$ from the environment
and monitors the cost signal $c$, which 
will become positive ($c>0$) when encountering safety violation. 
The trajectory will be terminated with safety violation.
Without safety violation, the next state reached from $s_t$ will be $s_{t'}=s_{t+H}$.
The skill-based transition data,
$D=\{(s_t, \zvec_t, \tilde{r}_t, s_{t'})\}$, 
are collected from the online interactions to train the safe skill policy $\pi_\theta(\cdot)$.
Meanwhile the state-skill decision pairs are collected as PU examples
in a similar way as on the demonstrations, such that
\begin{align}
	\!\!\!
 \mathcal{P}=\left\{(s_t,\zvec_t), (s_i,\zvec\!\sim\! q_\psi(\cdot|s_i))|i\in \{t\!+\!1\!:t'-1\}\right\}.
\label{dpairs}
\end{align}
These are then integrated with the existing PU data to continuously adapt 
the skill risk predictor to the online environment in {\em real-time},
enhancing and accelerating the online safe RL policy learning.
The full procedure of the proposed online safe RL learning is presented in Algorithm~\ref{algorithm:policy}.

In this work, we deploy a skill-based 
Soft Actor-Critic (SAC) algorithm~\cite{haarnoja2018soft} to learn the skill policy network $\pi_\theta(\cdot)$
on the collected data $D$,
which enforces behavior cloning
by replacing the entropy regularizer in the optimization objective of SAC with a 
KL-divergence regularizer, 
$KL(\pi_\theta(\cdot|s),q_\psi(\cdot|s))$,
between the skill policy network  $\pi_\theta(\cdot|s)$
and the pre-trained prior network $q_\psi(\cdot|s)$
\cite{pertsch2021accelerating}.
%
%%%%%%%%%%%%%%%%%%%%%%%%%%%%%%%%%%%%%%%
\begin{algorithm}[t]
\caption{Online Safe Policy Learning}
\label{algorithm:policy}
\textbf{Input:} 
skill prior $q_\psi(\cdot|s)$, decoder $p_\nu(\cdot|\zvec)$,\\ 
$\qquad$ skill risk predictor $P_\zeta(c|s,\zvec)$, $D^p$ and $D^u$ \\
\textbf{Initialize:}
data buffer $D$, skill policy network $\pi_\theta(\zvec|s)$\\
\textbf{Procedure:}
\begin{algorithmic}[1]
\FOR{each episode}
\STATE Randomly start from a state $s_0$, set $t=0$
\FOR{every $H$ environment steps}
	\STATE ${\zvec}_t\leftarrow \text{Risk\_Planning}(\pi_\theta(\cdot|s_t),P_\zeta(c|s_t,\zvec))$
	\STATE Sample $\textbf{a}_t=a_{t:t+H-1}$ from decoder $p_\nu(\cdot|{\zvec}_t)$
    \STATE Execute $\textbf{a}_t$: stop current trajectory $\tau$ when $c$$>$$0$
	\STATE Collect reward $\tilde{r}_t$ and get next state $s_{t'}$
	\STATE Add $\{s_t,{\zvec}_t,\tilde{r}_t, s_{t'}\}$ to $D$: ($t'$= $t$+$\min(H,|\tau|)$)
	\STATE Collect decision pairs $\mathcal{P}$ as in Eq.(\ref{dpairs})
	\STATE{\bf If} $c>0$ {\bf then: } Add $\mathcal{P}$ to $D^p$
	\STATE {\bf else: } Add $\mathcal{P}$ to $D^u$
	\STATE {\bf end if}
    \IF{$c > 0$ or reached max episode-steps}
     \STATE break out
    \ENDIF	
    \STATE $t=t'$
\ENDFOR
\STATE Update predictor $P_\zeta(c|s,\zvec)$ by minimizing Eq.~(\ref{eqa:pu_loss}) 
\STATE Update policy network $\theta$ following the skill-based SAC method on $D$.
\ENDFOR
\end{algorithmic}
\end{algorithm}
%%%%%%%%%%%%%%%%%%%%%%%%%%%%%%%%%%%%%%%

%
\begin{figure*}[th!]
\centering
\setlength{\abovecaptionskip}{0.cm}
\includegraphics[width=.85\textwidth,height=.7in]{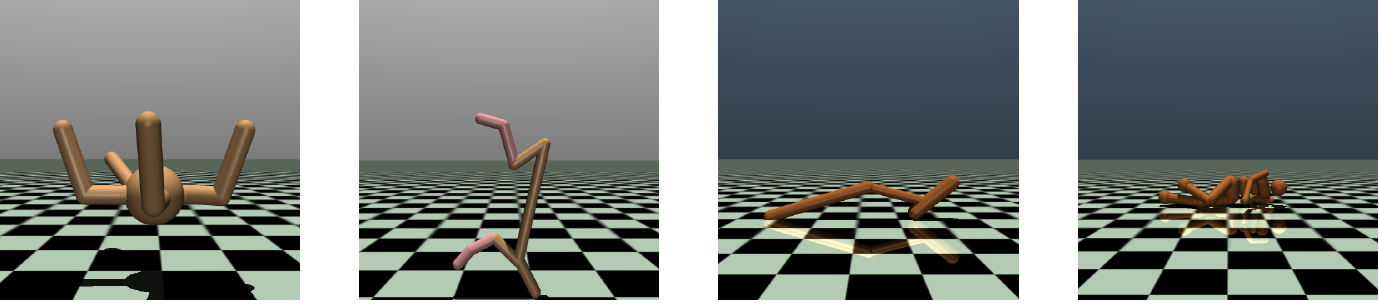}
\vskip -.05in
\caption{The four environments employed in the experiments are displayed from left to right:
        {\em Ant, Cheetah, Hopper, Humanoid}. The figures present instances of failure in each environment
        where safety constraints are violated.
	}
\label{fig:environments}
\end{figure*}

\section{Experiment}
\subsection{Experimental Settings}
\paragraph{\bf RL Environments}
We conducted experiments with
four benchmark robotic simulation environments, namely, {\em Ant, Cheetah, Hopper,} and {\em Humanoid},
utilizing a customized variant of the MuJoCo physics simulator~\cite{todorov2012mujoco} as 
introduced 
by~\citet{thomas2021safe}. 
In these environments, 
the RL agent halts 
upon encountering a safety violation.
In the {\em Ant} and {\em Hopper} environments, a safety violation occurs when the robot topples over.
In the {\em Cheetah} environment, a violation takes place when the robot's head hits the ground.
In the {\em Humanoid} environment, the human-like robot violates the safety constraint when its head falls to the
ground.
Figure~\ref{fig:environments} presents some instances of failure in these environments. 
The RL agent is trained to maximize cumulative rewards while adhering to the safety constraint. 
%%%%%%%%%%%%%%%%%%%%%%%%%%%%%%%%%%%%%%%%%%%%%%%
\paragraph{\bf Comparison Methods}
We compare our proposed SSkP approach with three state-of-the-art safe RL methods: 
CPQ~\cite{xu2022constraints}, SMBPO~\cite{thomas2021safe},
and Recovery RL~\cite{thananjeyan2021recovery}.
CPQ is a constraints penalized Q-learning method.
It learns from offline demonstration data, and 
penalizes the Bellman operator during policy training when encountering unsafe states. 
SMBPO is a model-based method 
that relies on an ensemble of Gaussian dynamics-based transition models. 
It penalizes trajectories that lead to unsafe conditions and avoids unsafe states under specific assumptions.
Recovery RL first learns a recovery policy from the offline demonstration data with the objective of
 minimizing safety violations. 
During online training, 
 the agent takes actions to maximize the reward signal in safe situations
 and falls back on the recovery policy to reduce safety violations if necessary.
\paragraph{\bf Implementation Details}
A fixed horizon length $H=10$ for skill action sequences is used in the experiments. 
The dimension of the skill vectors is set as 10. 
The PU risk predictor is implemented as a 3-layer MLP. Following prior work on PU
learning~\cite{xu2021positive}, the slack variable $\xi$ is set to 0.
For risk planning, we used $N_s=512$, $k=64$, and $N_p=6$. 
For comparison, we used the official implementations of Recovery RL~\cite{thananjeyan2021recovery}
and SMBPO~\cite{thomas2021safe}. The implementation of CPQ is adapted from the Offline Safe Reinforcement Learning
(OSRL) repository~\cite{liu2023datasets}. 
In the case of Recovery RL, both offline and online components are enabled.
As for CPQ~\cite{xu2022constraints}, the agent is pre-trained on the same offline dataset we collected and then trained
in the same manner in online environments. 
All results are collected over a total of $10^6$ online timesteps.

\subsection{Experimental Results}
\label{sec:experiment}
%
%%%%%%%%%%%%%%%%%%%%%%%%%%
\begin{figure*}[t]
\centering
    {\includegraphics[width=0.25\textwidth]{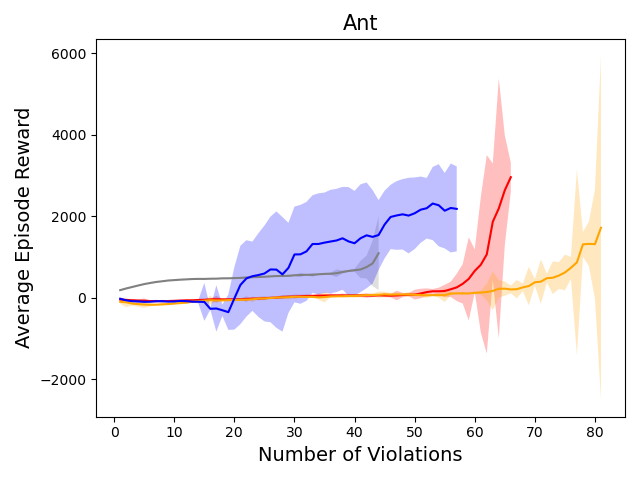}} 
  \hskip -.066in
{\includegraphics[width=0.25\textwidth]{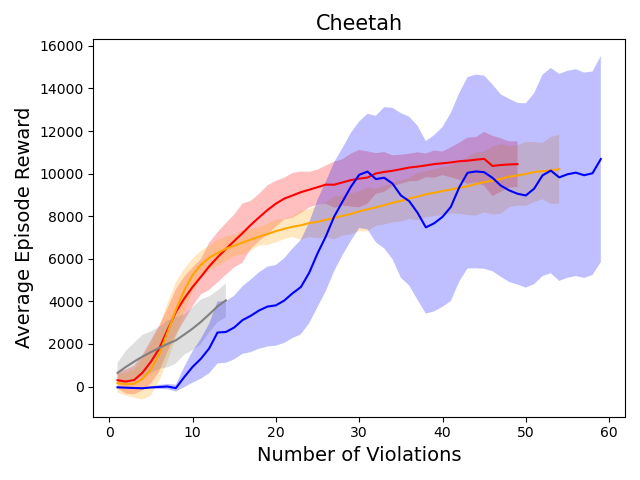}} 
 \hskip -.066in
	{\includegraphics[width=0.25\textwidth]{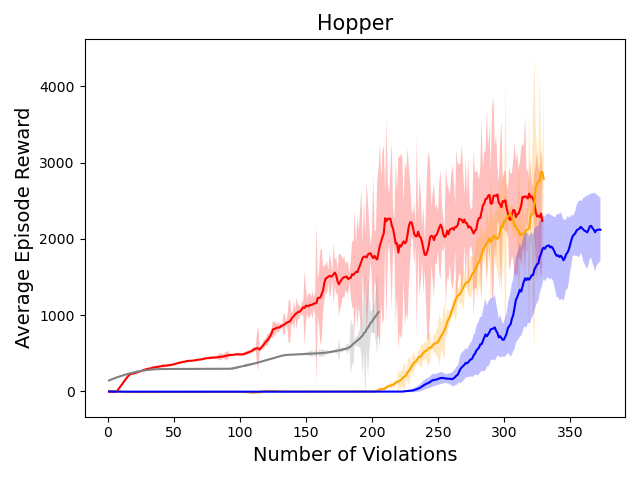}}
 \hskip -.066in
	{\includegraphics[width=0.25\textwidth]{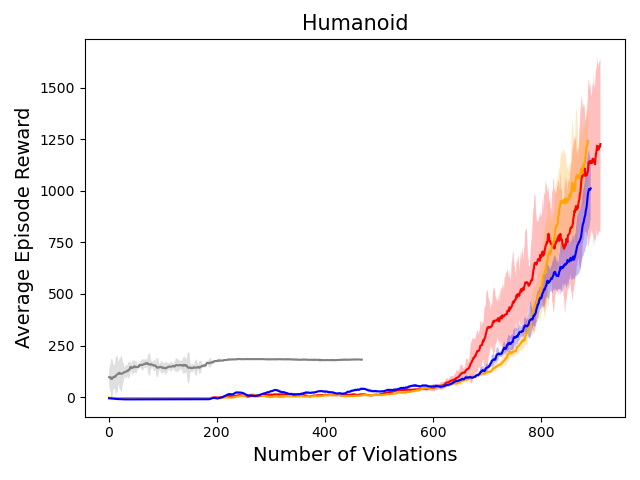}}
	{\includegraphics[width=0.4\textwidth]{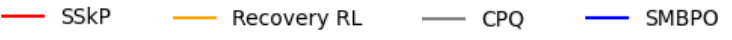}}
\vskip -.1in
\caption{
This figure presents the performance of each comparison method
in terms of the average episode reward vs. the total number of safety violations 
encountered during online training 
within a fixed total number of timesteps
	on all four environments:
{\em Ant}, {\em Cheetah}, {\em Hopper}, and {\em Humanoid}. 
The results represent the averages over three runs, with the shadow indicating the standard deviations.}
\label{fig:sskp}
\vskip -.15in
\end{figure*}
%%%%%%%%%%%%%%%%%%%%%%%%%%%%%
%
The comparison results for our proposed SSkP method and the other three safe RL methods 
in four robotic simulation environments are 
presented in Figure~\ref{fig:sskp}.
We used a similar evaluation strategy as the one in~\cite{thomas2021safe}. 
The results for all the methods are collected over the same total of $10^6$ online timesteps.
As the goal is to maximize the expected reward while minimizing the safety violation costs, 
we present the performance of each method in terms of its average episode reward versus 
the total number of safety violations encountered. 
Specifically, the x-axis depicts the cumulative safety violations encountered by the RL agent 
throughout the entire online training process,
while the y-axis reflects the average episode rewards 
with the increasing of numbers of violations.
These plots effectively illustrate the trade-off between reward maximization and 
risk (safety violation) minimization. 
A higher average episode reward with the same number of safety violations indicates better performance
in policy learning with the same cost. 

We can see that across all four environments, CPQ exhibits an initial advantage with a higher 
starting point and eventually halts with a very low average episode reward.  
This demonstrates that CPQ {\em failed to learn} a good policy function within the total $10^6$ online timesteps.
Although it only encountered a lower total number of violations,
the inability to effectively perform RL failed the ultimate goal. 
This can be attributed to that CPQ pre-trains its policy on the offline demonstration dataset. 
In contrast, both our proposed SSkP and Recovery RL do not rely on policy learning from offline demonstrations.
SSkP learns the skill model and the skill risk predictor from the offline demonstration data
and deploys them to support the online safe RL policy learning. 
SSkP outperforms Recovery RL in all four environments,
producing much higher average rewards with lower numbers of safety violations. 
SSkP also largely outperforms SMBPO in a similar way 
in three out of the four environments, except for the {\em Ant} environment; 
in {\em Ant}, SMBPO demonstrates a similar inability as CPQ in terms of 
learning a good policy to maximize the expected reward. 
Overall, the proposed SSkP method produces the most effective performance 
in all the four environments, outperforming the other comparison methods.
This validates the effectiveness of SSkP
for advancing safe RL by 
exploiting offline demonstration data.

%%%%%%%%%%%%%%%%%%%%%%%%%
\begin{table}[t]
\centering
\caption{
The table presents comparison results in all the four environments 
in terms of the ratio between Per-timestep Reward (PtR) and \#Violations (PtR/\#V ($\times 10^{3}$)).
This metric reflects the cost-sensitive sample efficiency of the online safe RL method. 
The results are averages over three runs.\\
}
\label{table:timestep_nostd}
\setlength{\tabcolsep}{6pt}	
\resizebox{1\columnwidth}{!}
{
\begin{tabular}{l|c|c|c|c}
\Xhline{1pt}
& \multicolumn{1}{c|}{Ant} & \multicolumn{1}{c|}{Cheetah} 
& \multicolumn{1}{c|}{Hopper} & \multicolumn{1}{c}{Humanoid} \\
\hline
SSkP        & $23.54$         & $\mathbf{173.72}$ 
            & $\mathbf{8.86}$ & $\mathbf{0.72}$    \\
Recovery RL & $13.12$         & $146.30$
            & $7.10$          & $0.71$             \\
CPQ         & $11.80$         & $92.25$
            & $6.21$          & $0.38$             \\
SMBPO       & $\mathbf{28.68}$& $147.00$
            & $5.74$          & $0.69$             \\
\Xhline{1pt}
\end{tabular}%
}
\vskip -.15in
\end{table}

%
%%%%%%%%%%
To provide a quantitative measure for the performance of an online safe RL agent
throughout the entire online learning process, 
we further introduce a new metric to compute the ratio
between the Per-timestep Reward (PtR) and the total number of safety Violations (\#V),
denoted as {\em PtR/\#V}. 
PtR is calculated by dividing the
cumulative episode reward across the entire online training duration 
by the total number of timesteps, which indicates the sample efficiency of the RL agent. 
Specifically, let $E$ represent the total number of episodes, $R_e$ denote the episode reward at episode $e$,
$T$ denote the total number of timesteps.
Then PtR is computed as ${\sum_{e=1}^E R_e/T}$. 
By further computing the ratio between PtR and the total number of safety violations, 
{\em PtR/\#V} takes the safety into consideration 
and can be used as a cost-sensitive sample efficiency metric for safe RL, 
which can capture the tradeoff between the learning efficiency of the safe RL agent
and the cost of encountering safety violations. 
The objective of safe RL is to maximize the reward while minimizing safety violation costs, 
naturally favoring a larger {\em PtR/\#V} ratio value.
%

%%%%%%%%%%%%%%%%%%%%%%%%
\begin{figure}[t!]
\centering
 {\includegraphics[width=0.243\textwidth]{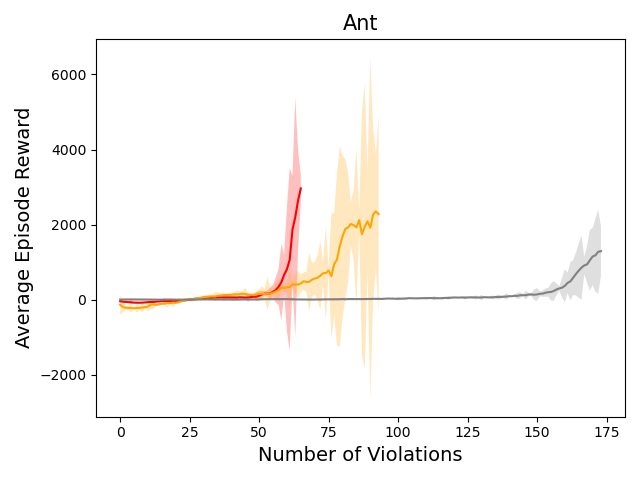}} 
\hskip -.067in	{\includegraphics[width=0.243\textwidth]{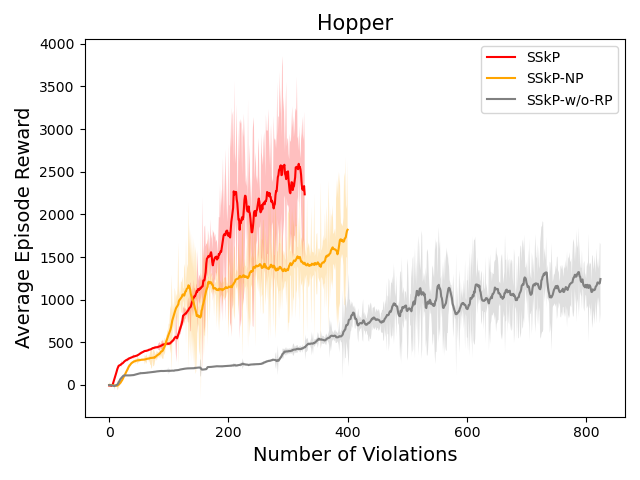}}
\vskip -.05in
\caption{
The ablation study results
in two environments: {\em Ant} and {\em Hopper} by comparing three methods: 
SSkP---the proposed approach; SSkP-NP---the variant that replaces risk planning
with a naive planning process; and SSkP-w/o-RP---the variant that drops risk predictor and risk planning
from SSkP. 
Each plot displays the average reward vs. the total number of safety violations 
encountered during online training within a fixed total number of timesteps. 
The results are averages of three runs. 
	}
\label{fig:ablation}
\vskip -.12in
\end{figure}
%%%%%%%%%%%%%%%%%%%%%%%%%

We calculated the average {\em PtR/\#V} values over three runs 
for all the comparison methods 
in all the four experimental 
environments,
and reported the comparison results 
in Table~\ref{table:timestep_nostd}, 
where the {\em PtR/\#V} numbers are scaled at $10^{3}$ for clarity of presentation.
Notably, under the PtR/\#V metric, our SSkP method outperforms all the other comparison methods 
in three out of the total four environments,
except for the {\em Ant} environment, where SSkP produced the second-best result.
The comparison method, CPQ, that 
has been shown to fail to learn in the figures, produces 
poor PtR/\#V values in all the environments.
Particularly in {\em Cheetah} and {\em Hopper},  
SSkP produces notable performance gains over all the other methods.
These results validate the superior efficiency and efficacy of our SSkP 
for online safe RL.  

%%%%%%%%%%%%%%%%%%%%%%%%
%
\begin{figure*}[t!]
\centering
    {\includegraphics[width=0.22\textwidth]{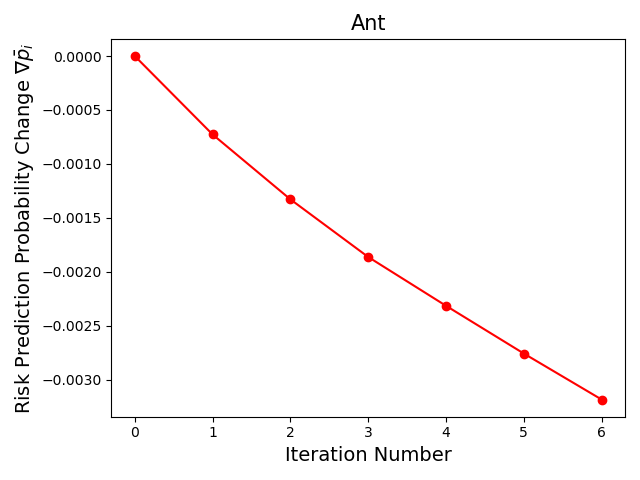}} 
	{\includegraphics[width=0.22\textwidth]{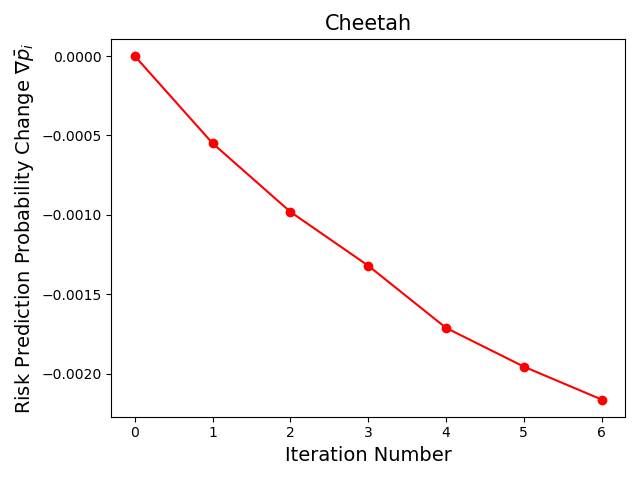}}
	{\includegraphics[width=0.22\textwidth]{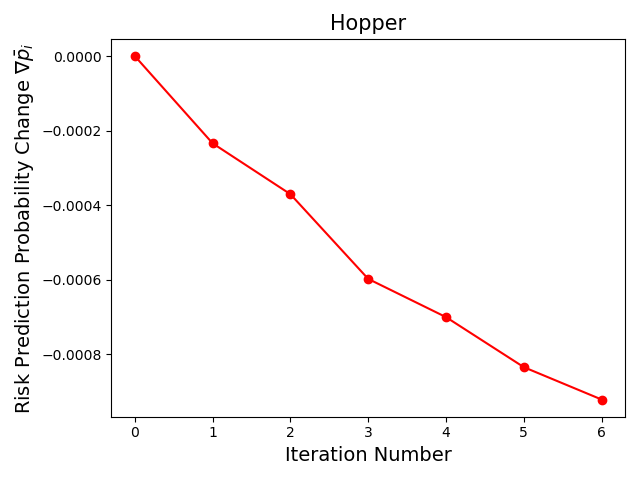}}
	{\includegraphics[width=0.22\textwidth]{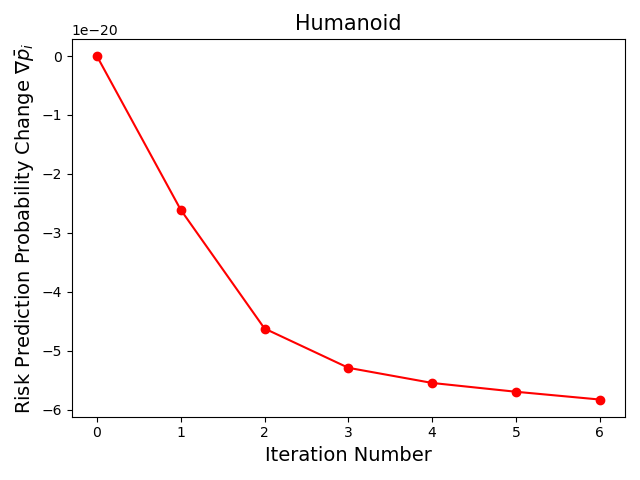}}
\vskip -.1in
\caption{
	Risk prediction probability changes, $\nabla\bar{p}_i=\bar{p}_i-\bar{p}_0$,  
along the planning iteration number $i$
	from the initial average risk prediction probability $\bar{p}_0$. 
The results are the averages computed with the risk planning procedure 
on 100 randomly sampled states $s_t$.
}
\label{fig:cem-prob}
\vskip -.05in
\end{figure*}
%%%%%%%%%%%%%%%%%%%%%%%%

%%%%%%%%%%%%%%%%%%%%
\subsection{Ablation Study}
The main contribution of the proposed SSkP approach lies in
devising two novel components: 
the {\em risk planning} component and the {\em skill risk predictor}.  
We conducted an ablation study to investigate their impact on
the performance of SSkP. 

The risk planning component in SSkP iteratively improves the safety of skills
by leveraging the skill risk predictor, 
aiming to generate and deploy the most effective safe skill decision. 
To investigate the extent to which the proposed risk planning process enhances safe policy learning performance,
we introduced an alternative {\em naive planning} baseline as a comparison. 
Naive planning samples $N_s$ skills using the current safe policy $\pi_\theta(\cdot|s_t)$
at the given state $s_t$, 
evaluates them using the current skill risk predictor,
and selects the best skill with the lowest predicted risk {\em in a single iteration}. 
We denote the variant of SSkP with naive planning instead of the proposed risk planning
as {\em SSkP-NP}.

The SSkP-NP variant nevertheless still leverages the skill risk predictor.  
To further investigate the impact of the skill risk predictor,
we introduced another variant, {\em SSkP-w/o-RP}, which drops the skill risk predictor
learning and deployment from both the offline and online learning stages. 
Consequently, risk planning that depends on skill risk assessment is also disabled
in the online RL stage,
while the skill decisions are produced directly by the skill policy function.

We compared the proposed full approach SSkP with 
the two variants, SSkP-NP and SSkP-w/o-RP, 
in the {\em Ant} and {\em Hopper} environments, and the experimental results are presented in Figure~\ref{fig:ablation}.
The curves in the figure reveal that our proposed SSkP with risk planning 
clearly outperforms the ablation variant SSkP-NP with naive planning
in both environments. 
In the {\em Hopper} environment, SSkP-NP exhibits a very brief faster improvement
during the early training stage but experiences a subsequent decline. 
Our proposed full approach SSkP produces a much better policy function 
that achieves substantially much higher average episode reward than SSkP-NP
with smaller cost---the number of safety violations. 
This validates the contribution put forth by the proposed risk planning process. 
We also note that 
by eliminating the skill risk predictor and consequently the entire risk planning, 
the variant SSkP-w/o-RP, while still leveraging the offline demonstration data through the skill model,
experiences a substantial performance decline compared to SSkP-NP. 
The results validate the significant contribution of the proposed skill risk prediction methodology,
which is the foundation of the proposed safe RL method SSkP. 

%%%%%%%%%%%%%%%%%%%%%%%%
\subsection{Further Study of Risk Planning Process}
The ablation study above validated the contribution of the proposed risk planning procedure
towards our overall safe RL approach, SSkP. 
In this subsection, we further study 
the efficacy of the risk planning procedure in Algorithm~\ref{algorithm:risk_planning}
as a zeroth-order solver for the non-convex optimization problem of 
$\argmin_{\zvec} P_\zeta(c=1|s_t, \zvec)$ 
by presenting the changes in the predicted risk probabilities of 
the sampled skills along the Gaussian distribution refinement iterations.

Specifically, in each experimental environment, 
given the trained risk predictor $P_\zeta(\cdot)$
and a sampled state $s_t$, 
we conduct risk planning with $N_p=6$ refinement iterations. 
From each Gaussian distribution $\mathcal{N}(\muvec_i,\text{diag}(\sigmavec_i^{2}))$,
along the iterations $i\in\{0, 1, \cdots, N_p\}$, 
we sample $N_s$ skills $\{\zvec_i^j\}_{j=1}^{N_s}$ 
and calculate the average of their predicted risk probabilities, 
$\bar{p}_i = \frac{1}{N_s}\sum_j^{N_s} P_\zeta(c=1|s_t,\zvec_i^j)$. 
To emphasize the effect of reducing risks of the sampled skills, 
we report the changes of the average risk probability 
from the initial iteration $0$; i.e., we record
$\nabla\bar{p}_i = \bar{p}_i-\bar{p}_0$ for each iteration $i$. 
We repeat this risk planning process over 100 randomly sampled
states $\{s_t\}$, and report the average results 
in Figure \ref{fig:cem-prob} for all the four experimental environments.
We can see with the increase of the risk planning iterations, 
$-\nabla\bar{p}_i$ becomes larger and hence $\bar{p}_i$ becomes smaller,  
indicating the sampled skills from each current Gaussian distribution 
are safer than previous iterations. 
Overall, the results validate that the risk planning process 
can effectively find safer skills $\zvec$ 
by minimizing $P_\zeta(c=1|s_t, \zvec)$.  

%%%%%%%%%%%%%%%%%%%%%%%%%%%%%%%%%%%%%%%%%%%%%%%%%
\section{Conclusion}
In this paper, we introduced a Safe Skill Planning (SSkP) method to 
address the challenge of online safe RL 
by effectively exploiting a prior demonstration dataset.
First, we deployed a deep skill model to extract safe behavior patterns from the demonstrations and   
proposed a novel skill risk predictor for decision safety evaluation,
which is trained through 
PU learning over the state-skill pairs. 
Second, by leveraging the risk predictor, 
we devised a new and simple risk planning process to iteratively identify reliable safe skill decisions
in online RL environments and support online safe RL policy learning.  
We compared the proposed method with several state-of-the-art safe RL methods 
in four benchmark robotic simulation environments. 
The experimental results demonstrate that our method yields notable improvements
over previous online safe RL approaches.

%%%%%%%%%%%%%%%%%%%%%%%%%%%%%%%%%%%%%%%%%%%%
\bibliography{icml2024}
\bibliographystyle{icml2024}

%%%%%%%%%%%%%%%%%%%%%%%%%%%%%%%%%%%%%%%%%%%%%%%%%%%%%%%%%%%%%%%%%%%%%%%%%%%%%%%
%%%%%%%%%%%%%%%%%%%%%%%%%%%%%%%%%%%%%%%%%%%%%%%%%%%%%%%%%%%%%%%%%%%%%%%%%%%%%%%
% APPENDIX
%%%%%%%%%%%%%%%%%%%%%%%%%%%%%%%%%%%%%%%%%%%%%%%%%%%%%%%%%%%%%%%%%%%%%%%%%%%%%%%
%%%%%%%%%%%%%%%%%%%%%%%%%%%%%%%%%%%%%%%%%%%%%%%%%%%%%%%%%%%%%%%%%%%%%%%%%%%%%%%
\newpage
\appendix
\onecolumn

\section{Alternative Evaluation of Experimental Results}
\begin{figure*}[th!]
\centering
{\includegraphics[width=0.25\textwidth]{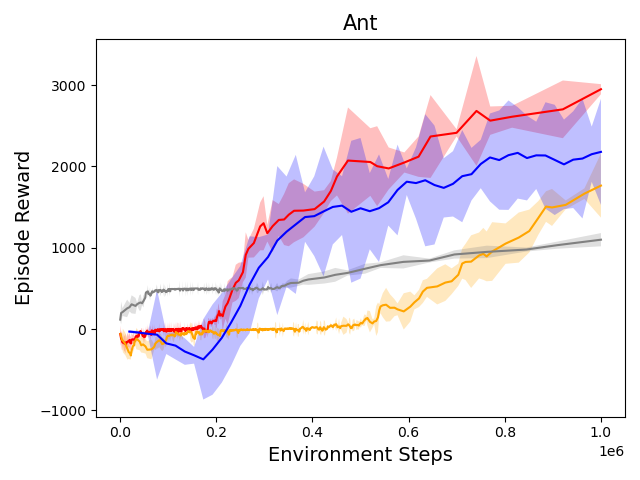}} 
\hskip -.066in
{\includegraphics[width=0.25\textwidth]{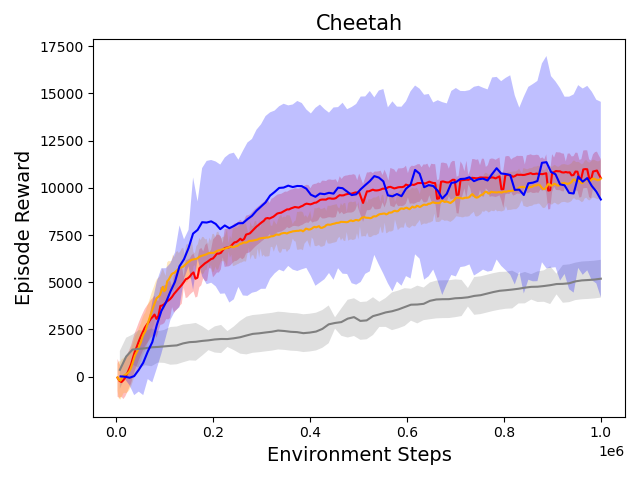}}
\hskip -.066in
{\includegraphics[width=0.25\textwidth]{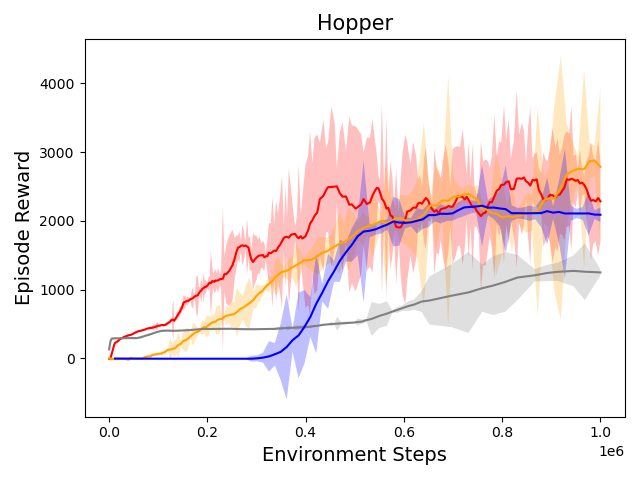}}
\hskip -.066in
{\includegraphics[width=0.25\textwidth]{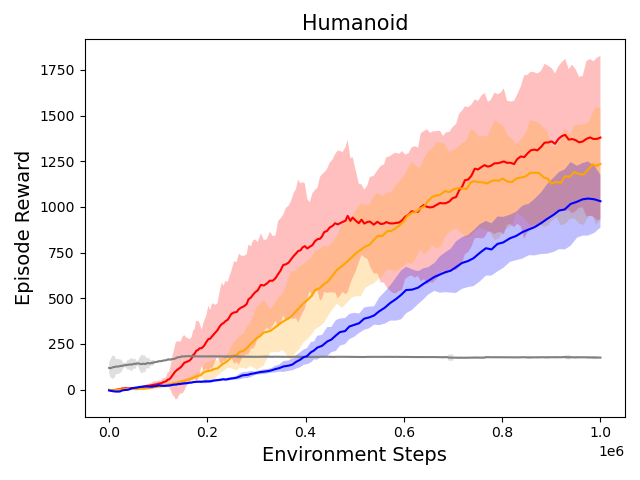}}
{\includegraphics[width=0.25\textwidth]{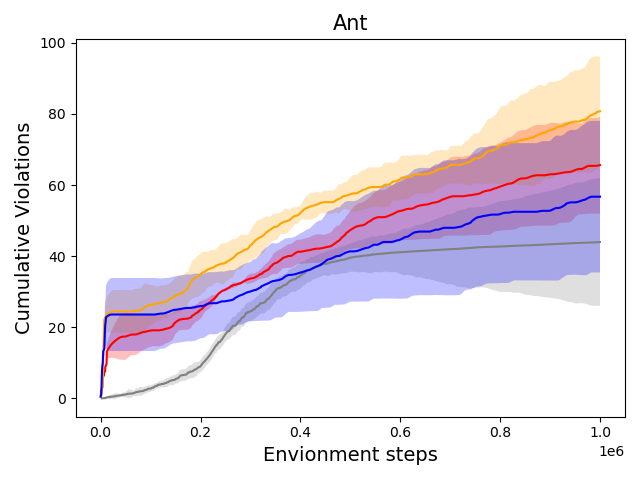}} 
\hskip -.066in
{\includegraphics[width=0.25\textwidth]{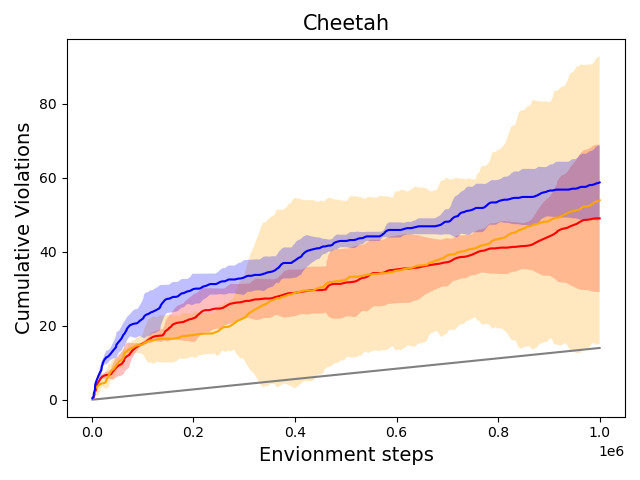}}
\hskip -.066in
{\includegraphics[width=0.25\textwidth]{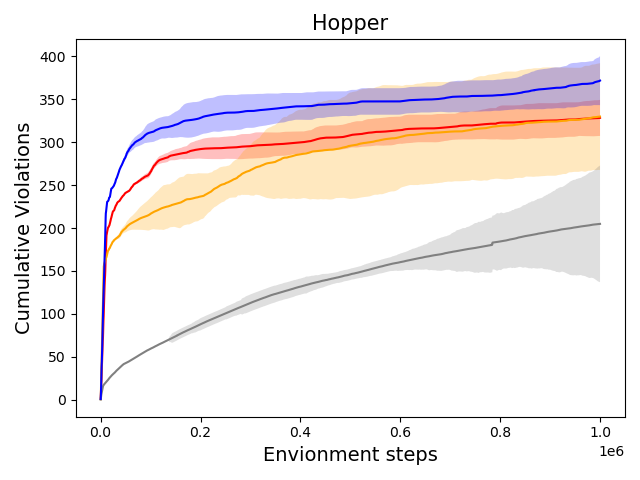}}
\hskip -.066in
{\includegraphics[width=0.25\textwidth]{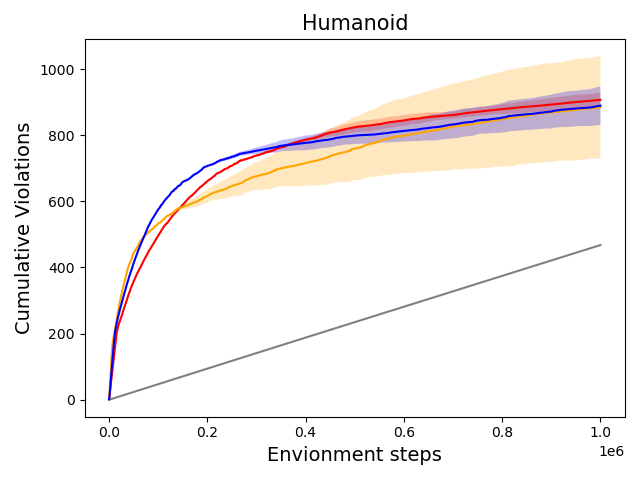}}
{\includegraphics[width=0.4\textwidth]{figure/exp_legend.pdf}}
\vskip -.1in
\caption{
The alternative evaluation of the safe RL results presents 
the episode rewards and 
the cumulative number of violations separately along the environment steps. 
{\bf Top:}
Sample efficiency curves illustrating 
episode rewards 
v.s. the total number of environmental steps across four environments.
{\bf Bottom:}
Violation curves illustrating the total number of violations v.s. the total number of environmental steps across four environments. The results are 
averages of three runs.
}
\label{fig:sample_efficiency}
\vskip -.1in
\end{figure*}

We have introduced an alternative evaluation of our experimental results in Section \ref{sec:experiment}, 
simultaneously presenting sample efficiency curves and violation curves. 
This approach offers an intuitive understanding of the overall performance of our safe RL agent,
illustrating performance and safety metrics across environmental steps.
The results are illustrated in Figure \ref{fig:sample_efficiency}.
Notably, on the {\em Ant}, {\em Hopper}, and {\em Humanoid} environments, our SSkP demonstrates superior performance
based on sample efficiency curves, while on the {\em Cheetah} environments, SSkP exhibits comparable performance to
Recovery RL and SMBPO. These findings highlight SSkP's robust performance across environments,
even in the absence of explicit safety constraints.
Although CPQ displays the lowest cumulative violations compared to other methods, 
it fails to achieve acceptable episode rewards, indicating its incapacity to learn an effective policy while following safety constraints.
For the {\em Cheetah}, {\em Hopper}, and {\em Humanoid} environments, as the number of environmental steps increases, 
SSkP exhibits comparable safety violations with the second-best comparison method (excluding CPQ), 
while outperforming comparison methods in terms of episode rewards.

\end{document}